\documentclass[10pt,twocolumn,letterpaper]{article}

\usepackage{cvpr}
\usepackage{times}
\usepackage{epsfig}
\usepackage{graphicx}
\usepackage{amsmath}
\usepackage{amssymb}
\usepackage{color}
\usepackage{amsmath}
\usepackage{amssymb}
\usepackage{fancyhdr}
\usepackage{listings}
\usepackage{multirow}
\usepackage{graphicx}
\usepackage{verbatim}
\usepackage{bbm}
\usepackage{color}
\usepackage{mathtools}
\usepackage{subfigure}
\usepackage{pifont}
\usepackage{makecell}
\usepackage{multirow}
\usepackage{booktabs}
\usepackage{float}
\usepackage[x11names, rgb]{xcolor}
\usepackage{amsmath,amsfonts,amsthm,bm}
\usepackage{enumitem}
\usepackage[breaklinks=true,bookmarks=false]{hyperref}

\DeclareMathOperator*{\argmax}{argmax\;}
\DeclareMathOperator*{\argmin}{argmin\;}

\cvprfinalcopy % *** Uncomment this line for the final submission

\def\cvprPaperID{6169} % *** Enter the CVPR Paper ID here

\definecolor{dgreen}{rgb}{0.0, 0.5, 0.0}
\definecolor{bured}{rgb}{0.8, 0.0, 0.0}
\newcommand{\rulesep}{\unskip\ \vrule\ }
\ifcvprfinal\fi
\begin{document}

%%%%%%%%% TITLE
\title{Characterizing and Avoiding Negative Transfer}

\author{
Zirui Wang, Zihang Dai, Barnab\'{a}s P\'{o}czos, Jaime Carbonell \\
Cargenie Mellon University \\
{\tt\small \{ziruiw,dzihang,bapoczos,jgc\}@cs.cmu.edu}
}

\maketitle
%\thispagestyle{empty}

%%%%%%%%% ABSTRACT
\begin{abstract}
% Top-performing machine learning algorithms are usually trained on massive amount of labeled data.
When labeled data is scarce for a specific target task, transfer learning often offers an effective solution by utilizing data from a related source task. However, when transferring knowledge from a less related source, it may inversely hurt the target performance, a phenomenon known as negative transfer. Despite its pervasiveness, negative transfer is usually described in an informal manner, lacking rigorous definition, careful analysis, or systematic treatment. This paper proposes a formal definition of negative transfer and analyzes three important aspects thereof. Stemming from this analysis, a novel technique is proposed to circumvent negative transfer by filtering out unrelated source data. Based on adversarial networks, the technique is highly generic and can be applied to a wide range of transfer learning algorithms. The proposed approach is evaluated on six state-of-the-art deep transfer methods via experiments on four benchmark datasets with varying levels of difficulty. Empirically, the proposed method consistently improves the performance of all baseline methods and largely avoids negative transfer, even when the source data is degenerate.
\end{abstract}
\vspace{-0.2in}

%%%%%%%%% BODY TEXT
\section{Introduction}
The development of deep neural networks (DNNs) has improved the state-of-the-art performance on a wide range of machine learning problems and applications. However, DNNs often require a large amount of labeled data to train well-generalized models and as more classical methods, DNNs rely on the assumption that training data and test data are drawn from the same underlying distribution. 
In some cases, collecting large volumes of labeled training data is expensive or even prohibitive. 
Transfer learning~\cite{pan2010survey} addresses this challenge of data scarcity by utilizing previously-labeled data from one or more source tasks. 
The hope is that this source domain is related to the target domain and thus transferring knowledge from the source can improve the performance within the target domain. 
This powerful paradigm has been studied under various settings \cite{weiss2016survey} and has been proven effective in a wide range of applications \cite{zamir2018taskonomy, long2015learning, moon2017completely}. 

However, the success of transfer learning is not always guaranteed. 
If the source and target domains are not sufficiently similar, transferring from such weakly related source may hinder the performance in the target, a phenomenon known as negative transfer. 
The notion of negative transfer has been well recognized within the transfer learning community \cite{pan2010survey,weiss2016survey}. 
An early paper \cite{rosenstein2005transfer} has conducted empirical study on a simple binary classification problem to demonstrate the existence of negative transfer. 
Some more recent work \cite{duan2012exploiting,ge2014handling,cao2017partial} has also observed similar negative impact while performing transfer learning on more complex tasks under different settings. 

Despite these empirical observations, little research work has been published to analyze or predict negative transfer, and the following questions still remain open: First, while the notion being quite intuitive, it is not clear how negative transfer should be defined exactly. For example, how should we measure it at test time? What type of baseline should we compare with? Second, it is also unknown what factors cause negative transfer, and how to exploit them to determine that negative transfer may occur. Although the divergence between the source and target domain is certainly crucial, we do know how large it must be for negative transfer to occur, nor if it is the only factor. Third and most importantly, given limited or no labeled target data, how to detect and/or avoid negative transfer.

In this work, we take a step towards addressing these questions. 
We first derive a formal definition of negative transfer that is general and tractable in practice. Here tractable means we can explicitly measure its effect given the testing data. 
This definition further reveals three underlying factors of negative transfer that give us insights on when it could occur. 
Motivated by these theoretical observations, we develop a novel and highly generic technique based on adversarial networks to combat negative transfer. 
In our approach, a discriminator estimating both marginal and joint distributions is used as a gate to filter potentially harmful source data by 
reducing the bias between source and target risks, which corresponds to the idea of importance reweighting~\cite{cortes2010learning,yu2012analysis}.
Our experiments involving eight transfer learning methods and four benchmark datasets reveal the three factors of negative transfer. In addition, we apply our method to six state-of-the-art deep methods and compare their performance, demonstrating that our approach substantially improves the performance of all base methods under potential negative transfer conditions by largely avoiding negative transfer.

\section{Related Work}
\textit{Transfer learning} \cite{pan2010survey,Yang2012Transfer} uses knowledge learned in the source domain to assist training in the target domain.  Early methods exploit conventional statistical techniques such as instance weighting \cite{huang2007correcting} and feature mapping \cite{pan2011domain,Uguroglu2011feature}. Compared to these earlier approaches, deep transfer networks achieve better results in discovering domain invariant factors \cite{yosinski2014transferable}. Some deep methods \cite{long2015learning,sun2016deep} transfer via distribution (mis)match measurements such as Maximum Mean Discrepancy (MMD) \cite{huang2007correcting}. More recent work\cite{ganin2016domain,tzeng2015simultaneous,cao2017partial,sankaranarayanan2017generate} exploit generative adversarial networks (GANs) \cite{goodfellow2014generative} and add a subnetwork as a domain discriminator. These methods achieve state-of-the-art on computer vision tasks \cite{sankaranarayanan2017generate} and some natural language processing tasks \cite{moon2017completely}. However, none of these techniques are specifically designed to tackle the problem of negative transfer.

\textit{Negative transfer}  Early work that noted negative transfer \cite{rosenstein2005transfer} was targeted at simple classifiers such as hierarchical Naive Bayes. Later, similar negative effects have also been observed in various settings including multi-source transfer learning \cite{duan2012exploiting}, imbalanced distributions \cite{ge2014handling} and partial transfer learning \cite{cao2017partial}. While the importance of detecting and avoiding negative transfer has raised increasing attention \cite{weiss2016survey}, the literature lacks in-depth analysis.

\section{Rethink Negative Transfer}
\label{sec:rethink}

\paragraph{Notation.} We will use $P_S(X, Y)$ and $P_T(X, Y)$, respectively, to denote the the joint distribution in the source and the target domain, where $X$ is the input random variable and $Y$ the output.
Following the convention, we assume having access to labeled source set $\mathcal{S} = \{(x_s^i, y_s^i)\}_{i=1}^{n_s}$ sampled from the source joint $P_S(X, Y)$, a labeled target set $\mathcal{T}_l = \{(x_l^j, y_l^j)\}_{j=1}^{n_l}$ drawn from the target joint $P_T(X, Y)$, and an unlabeled target set $\mathcal{T}_u = \{x_u^k\}_{k=1}^{n_u}$ from the target marginal $P_T(X)$.
For convenience, we define $\mathcal{T} = (\mathcal{T}_l, \mathcal{T}_u)$.

\paragraph{Transfer Learning.} Under the notation, transfer learning aims at designing an algorithm $A$, which takes both the source and target domain data $\mathcal{S}, \mathcal{T}$ as input, and outputs a better hypothesis (model) $h = A(\mathcal{S}, \mathcal{T})$, compared to only using the target-domain data .
For model comparison, we will adapt the standard expected risk, which is defined as
\begin{equation}
    R_{P_T}(h) \coloneqq \mathbb{E}_{x, y \sim P_T} \left[ \ell(h(x), y) \right], 
\end{equation}
with $\ell$ being the specific task loss.
To make the setting meaningful, it is often assumed that $n_s \gg n_l$.

\paragraph{Negative Transfer.} The notion of negative transfer lacks a rigorous definition.
A widely accepted description of negative transfer~\cite{pan2010survey, weiss2016survey} is stated as ``\textit{transferring knowledge from the source can have a negative impact on the target learner}''.
While intuitive, this description conceals many critical factors underlying negative transfer, among which we stress the following three points:
\begin{enumerate}[leftmargin=*]
\item \textit{Negative transfer should be defined w.r.t. the algorithm}. 
Specifically, the informal description above does not specify what the negative impact is compared with. 
For example, it will be misleading to only compare with the best possible algorithm only using the target data, i.e., defining negative transfer as \vspace{-0.5em}
\begin{equation}
R_{P_T}(A(\mathcal{S},\mathcal{T})) > \min_{A'} R_{P_T}(A'(\emptyset,\mathcal{T})),\vspace{-0.5em}
\end{equation}
because the increase in risk may not come from using the source-domain data, but the difference in algorithms.
Therefore, to study negative transfer, one should focus on a specific algorithm at a time and compare its performance with and without the source-domain data. 
Hence, we define the \textit{negative transfer condition} (NTC)\footnote{More discussion in the supplementary.} for any algorithm $A$ as \vspace{-0.5em}
\begin{equation}
R_{P_T}(A(\mathcal{S},\mathcal{T})) > R_{P_T}(A(\emptyset,\mathcal{T})).\vspace{-0.5em}
\label{eq:ntc}
\end{equation}
For convenience, we also define the \textit{negative transfer gap} (NTG) as a quantifiable measure of
negative transfer: \vspace{-0.5em}
\begin{equation}
R_{P_T}(A(\mathcal{S},\mathcal{T})) - R_{P_T}(A(\emptyset,\mathcal{T})),\vspace{-0.5em}
\label{eq:ntg}
\end{equation}
and we say that negative transfer occurs if the negative transfer gap is positive and vice versa.

\item \textit{Divergence between the joint distributions is the root to negative transfer}.
As negative transfer is algorithm specific, it is natural to ask the question that whether there exists a transfer learning algorithm that can always improve the expected risk compared to its target-domain only baseline.
It turned out this depends on the divergence between $P_S(X, Y)$ and $P_T(X, Y)$ \cite{gong2016domain}. 
As an extreme example, assume $P_S(X) = P_T(X)$ and $P_S(Y \mid x)$ is uniform for any $x$. In the case, there is no meaningful knowledge in $P_S(X, Y)$ at all. 
Hence, exploiting $\mathcal{S} \sim P_S(X,Y)$ will almost surely harm the estimation of $P_T(Y \mid X)$, unless $P_T(Y \mid X)$ is uniform.

In practice, we usually deal with the case where there exists some ``systematic similarity'' between $P_S(X, Y)$ and $P_T(X, Y)$. 
Then, an ideal transfer would figure out and take advantage of the similar part, leading to improved performance. 
However, if an algorithm fails to discard the divergent part and instead rely on it, one can expect negative transfer to happen.
Thus, regardless of the algorithm choice, the distribution shift is the actual root to negative transfer.

\item \textit{Negative transfer largely depends on the size of the labeled target data}.
While the previous discussion focuses on the distribution level, an overlooked factor of negative transfer is the size of the labeled target data, which has a mixed effect.

On one hand, for the same algorithm and distribution divergence, NTC depends on how well the algorithm can do using target data alone, i.e. the RHS of Eq.\eqref{eq:ntc}. 
In zero-shot transfer learning\footnote{It is often referred to as unsupervised domain adaptation in literature.} \cite{ganin2015unsupervised,pei2018multi} where there is no labeled target data ($n_l = 0$), only using unlabeled target data would result in a weak random model and thus NTC is unlikely to be satisfied. 
When labeled target data is available \cite{rosenstein2005transfer,tzeng2015simultaneous,moon2017completely}, a better target-only baseline can be obtained using semi-supervised learning methods and so negative transfer is \textit{relatively} more likely to occur. 
At the other end of the spectrum, if there is an abundance of labeled target data, then transferring from a even slightly different source domain could hurt the generalization. 
Thus, this shows that negative transfer is \textit{relative}.

On the other hand, the amount of labeled target data has a direct effect on the feasibility and reliability of discovering shared regularity between the joint distributions. 
As discussed above, the key component of a transfer learning algorithm is to discover the similarity between the source joint $P_S(X, Y)$ and the target joint $P_T(X, Y)$. 
When labeled target data is not available ($n_l = 0$), one has to resort to the similarity between the marginals $P_S(X)$ and $P_T(X)$, which though has a theoretical limitation~\cite{ben2007analysis}. 
In contrast, if one has a considerable number of samples $(x_l, y_l) \sim P_T(X, Y)$ and $(x_s, y_s) \sim P_S(X, Y)$, the problem would be manageable.
Therefore, an ideal transfer learning algorithm may be able to utilize labeled target data to mitigate the negative impact of unrelated source information.
\end{enumerate}

With these points in mind, we next turn to the problem of how to avoid negative transfer in a systematic way. 

\section{Proposed Method}
As discussed in Section \ref{sec:rethink}, the key to achieving successful transfer and avoiding negative effects is to discover and exploit shared underlying structures between $P_S(X, Y)$ and $P_T(X, Y)$.
In practice, there are many possible regularities one may take advantage of. 
To motivate our proposed method, we first review an important line of work and show how the observation in section 3 helps us to identify the limitation.

\subsection{Domain Adversarial Network}
As a notable example, a recent line of work~\cite{long2015learning,ganin2015unsupervised,tzeng2017adversarial} has successfully utilized a domain-invariant feature space assumption to achieve knowledge transfer.
Specifically, it is assumed that there exists a feature space that is both shared by both source and target domains and discriminative enough for predicting the output.
By learning a feature extractor $F$ that can map both the source and target input to the same feature space, classifier learned on the source data can transfer to the target domain.

To find such a feature extractor, a representative solution is the Domain Adversarial Neural Network (DANN)~\cite{ganin2016domain}, which exploits a generative adversarial network (GAN) framework to train the feature extractor $F$ such that the feature distributions $P(F(X_S))$ and $P(F(X_T))$ cannot be distinguished by the discriminator $D$. 
Based on the shared feature space, a simple classifier $C$ is trained on both source and target data.
Formally, the objective can be written as:
\begin{align}
\label{eq:f1}
\argmin_{F, C} \argmax_{D} 
&\mathcal{L}_\text{CLF}(F, C) - \mu \mathcal{L}_\text{ADV}(F, D), \\
\label{eq:dann_clf_obj}
\mathcal{L}_\text{CLF}(F, C)
=&\, \mathbb{E}_{x_l, y_l \sim \mathcal{T}_L} \left[ \ell_\text{CLF}(C(F(x_l)), y_l) \right] \nonumber\\
+&\, \mathbb{E}_{x_s, y_s \sim \mathcal{S}} \left[ \ell_\text{CLF}(C(F(x_s)), y_s) \right], \\
\label{eq:dann_adv_obj}
\mathcal{L}_\text{ADV}(F, D)
=&\, \mathbb{E}_{x_u \sim P_T(X)} \left[ \log D(F(x_u)) \right] \nonumber\\
+&\, \mathbb{E}_{x_s \sim P_S(X)} \left[ \log (1 - D(F(x_s))) \right].
\end{align}
Intuitively, $\mathcal{L}_\text{CLF}$ is the supervised classification loss on both the target and source labeled data, $\mathcal{L}_\text{ADV}$ is the standard GAN loss treating $F(x_u)$ and $F(x_s)$ as the true and fake features respectively, and $\mu$ is a hyper-parameter balancing the two terms.
For more details and theoretical analysis, we refer readers to the original work~\cite{ganin2015unsupervised}.

Now, notice that the DANN objective implicitly makes the following assumption:
For any $x_s \in \mathcal{X}_s$, there exists a $x_t \in \mathcal{X}_t$ such that 
\[ 
 P_S(Y | x_s) = P_T(Y | x_t) = P(Y | F(x_s)) = P(Y | F(x_t)). 
\]
In other words, it is assumed that every single source sample can provide meaningful knowledge for transfer learning.
However, as we have discussed in Section \ref{sec:rethink}, some source samples may not be able to provide any knowledge at all. 
Consider the case where there is a source input $x_s \in \mathcal{X}_s$ such that $P_S(Y \mid x_s) \neq P_T(Y \mid x_t)$ for any $x_t$. 
Since $P(F(X_s)) = P(F(X_t))$ as a result of the GAN objective, there exists a $x' \in \mathcal{X}_t$ such that $F(x') = F(x_s)$ and hence $P(Y \mid F(x')) = P(Y \mid F(x_s))$.
Then, if $P(Y \mid F(x_s))$ is trained on the source data to match $P_S(Y \mid x_s)$, it follows 
\[ 
P(Y | F(x')) = P(Y | F(x_s)) = P(Y | x_s) \neq P(Y | x'). 
\]
As a result, relying on such ``unrelated'' source samples can hurt the performance, leading to negative transfer.
Motivated by this limitation, we next present a simple yet effective method to deal with harmful source samples in a systematic way. 

\begin{figure}[t]
\begin{center}
   \includegraphics[width=1.0\linewidth]{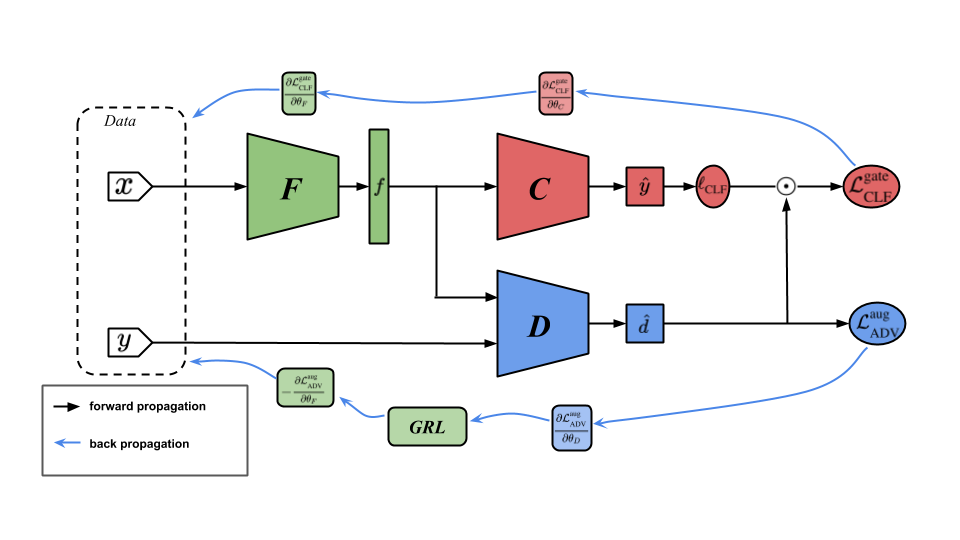}
\end{center}
   \caption{The architecture of proposed discriminator gate, where $f$ is the extracted feature layer, $\hat{y}$ and $\ell_\text{CLF}$ are predicted class label and its loss, $\hat{d}$ is the predicted domain label, $\mathcal{L}^\text{gate}_\text{CLF}$ is the classification loss, $\mathcal{L}^\text{aug}_\text{ADV}$ is the adversarial learning loss; GRL stands for Gradient Reversal Layer and $\odot$ is the Hadamard product.}
\label{fig:model}\vspace{-0.5em}
\end{figure}

\subsection{Discriminator Gate}
The limitation of DANN comes from the unnecessary assumption that all source samples are equally useful. 
To eliminate the weakness, a natural idea is to reweight each source sample in some proper manner.
To derive an appropriate weight, notice that the standard supervised learning objective can be rewritten as
\begin{equation}
\begin{aligned}
\mathcal{L}_\text{SUP} 
&= \mathbb{E}_{x, y \sim P_T(X, Y)} \left[ \ell_\text{CLF}(C(F(x)), y) \right] \\
&= \mathbb{E}_{x, y \sim P_S(X, Y)} \left[ \frac{P_T(x, y)}{P_S(x, y)} \ell_\text{CLF}(C(F(x)), y) \right]
\end{aligned}
\end{equation}
where the density ratio $\frac{P_T(x, y)}{P_S(x, y)}$ naturally acts as an importance weight~\cite{cortes2010learning,yu2012analysis} for the source data.
Hence, the problem reduces to the classic problem of density ratio estimation.

Here, we exploit a GAN discriminator to perform the density ratio estimation~\cite{uehara2016generative}. 
Specifically, the discriminator takes both $x$ and the paired $y$ as input, and try to classify whether the pair is from the source domain (fake) or the target domain (true). 
At any point, the optimal discriminator is given by $D(x, y) = \frac{P_T(x, y)}{P_T(x, y) + P_S(x, y)}$, which implies 
\[
\frac{P_T(x, y)}{P_S(x, y)} = \frac{D(x, y)}{1 - D(x, y)}.
\]
In our implementation, to save model parameters, we reuse the feature extractor to obtain the feature of $x$ and instantiate $D(x, y)$ as $D(F(x), y)$.
With the weight ratio, we modify the classification objective \eqref{eq:dann_clf_obj} in DANN as
\begin{equation}
\label{eq:gate_clf_obj}
\begin{aligned}
\mathcal{L}^\text{gate}_\text{CLF}(C, F) 
&= \mathbb{E}_{x_l, y_l \sim \mathcal{T}_L} \left[\ell_\text{CLF}(C(F(x_l)), y_l)\right] \\
+ \lambda &\mathbb{E}_{x_s, y_s \sim \mathcal{S}} \left[ \omega(x_s, y_s) \ell_\text{CLF}(C(F(x_s)), y_s) \right], \\
\omega(x_s, y_s) &= \text{SG}\left( \frac{D(x_s, y_s)}{1 - D(x_s, y_s)} \right)
\end{aligned}
\end{equation}
where $\text{SG}(\cdot)$ denotes stop gradient and $\lambda$ is another hyper-parameter introduce to scale the density ratio.
As the density ratio acts like a gating function, we will refer to mechanism as discriminator gate.

On the other hand, we also augment the adversarial learning objective \eqref{eq:dann_adv_obj} by incorporating terms for matching the joint distributions:
\begin{equation}
\label{eq:aug_adv_obj}
\begin{aligned}
&\mathcal{L}^\text{aug}_\text{ADV}(F, D)
= \mathbb{E}_{x_u \sim P_T(X)} \left[ \log D(F(x_u), \texttt{nil}) \right] \\
&\quad+ \mathbb{E}_{x_s \sim P_S(X)} \left[ \log (1 - D(F(x_s), \texttt{nil})) \right] \\
&\quad+ \mathbb{E}_{x_l, y_l \sim \mathcal{T}_L} \left[ \log D(F(x_l), y_l) \right] \\
&\quad+ \mathbb{E}_{x_s, y_s \sim \mathcal{S}} \left[ \log (1 - D(F(x_s), y_s)) \right],
\end{aligned}
\end{equation}
where $\texttt{nil}$ denotes a dummy label which does not provide any label information and it is included to enable the discriminator $D$ being used as both a marginal discriminator and a joint discriminator.
As a benefit, the joint discriminator can utilize unlabeled target data since labeled data could be scarce.
Similarly, under this objective, the feature network $F$ will receive gradient from both the marginal discriminator and the joint discriminator. 
Theoretically speaking, the joint matching objective subsumes the the marginal matching objective, as matched joint distribution implied matched marginals. 
However, in practice, the labeled target data $\mathcal{T}_L$ is usually limited, making the joint matching objective itself insufficient. 
This particular design choice echos our discussion about how the size of labeled target data can influence our algorithm design in Section \ref{sec:rethink}.

Combining the gated classification objective \eqref{eq:gate_clf_obj} and the augmented adversarial learning objective \eqref{eq:aug_adv_obj}, we arrive at our proposed approach to transfer learning
\begin{equation}
\argmin_{F, C} \argmax_{D} \mathcal{L}^\text{gate}_\text{CLF}(F, C) - \mu \mathcal{L}^\text{aug}_\text{ADV}(F, D).
\end{equation}
The overall architecture is illustrated in Figure \ref{fig:model}.
Finally, although the presentation of the proposed method is based on DANN, our method is highly general and can be applied directly to other adversarial transfer learning methods. 
In fact, we can even extend non-adversarial methods to achieve similar goals. 
In our experiments, we adapt six deep methods \cite{long2015learning,sun2016deep,ganin2015unsupervised,tzeng2017adversarial,cao2018partial,sankaranarayanan2017generate} of three different categories to demonstrate the effectiveness of our method.

\section{Experiments}
We conduct extensive experiments on four benchmark datasets to (1) analyze negative transfer and its three underlying aspects, and (2) evaluate our proposed discriminator gate on six state-of-the-art methods. 

\subsection{Datasets}
We use four standard datasets with different levels of difficulties: (1) small domain shift: Digits dataset, (2) moderate domain shift: Office-31 dataset, and (3) large domain shift: Office-Home and VisDA datasets.

\textbf{Digits} contains three standard digit classification datasets: MNIST, USPS, SVHN. Each dataset contains large amount of images belonging to 10 classes (0-9). This dataset is relatively easy due to its simple data distribution and therefore we only consider a harder case: \textbf{SVHN$\rightarrow$MNIST}. Specifically, SVHN \cite{netzer2011reading} contains 73K images cropped from house numbers in Google Street View images while MNIST \cite{lecun1998gradient} consists of 70K handwritten digits captured under constrained conditions.

\textbf{Office-31} \cite{saenko2010adapting} is the most widely used dataset for visual transfer learning. It contains 4,652 images of 31 categories from three domains: Amazon(\textbf{A}) which contains images from amazon.com, Webcam(\textbf{W}) and DSLR(\textbf{D}) which consist of images taken by web camera and SLR camera. We evaluate all methods across three tasks: \textbf{W$\rightarrow$D}, \textbf{A$\rightarrow$D}, and \textbf{D$\rightarrow$A}. We select these three settings because the other three possible cases yield similar results.

\textbf{Office-Home} \cite{venkateswara2017Deep} is a more challenging dataset that consists of about 15,500 images of 65 categories that crawled through several search engines and online image directories. In particular, it contains four domains: Artistic images(\textbf{Ar}), Clip Art(\textbf{Cl}), Product images(\textbf{Pr}) and Real-World images(\textbf{Rw}). We want to test on more interesting and practical transfer learning tasks involving adaptation from synthetic to real-world and thus we consider three transfer tasks: \textbf{Ar$\rightarrow$Rw}, \textbf{Cl$\rightarrow$Rw}, and \textbf{Pr$\rightarrow$Rw}. In addition, we choose to use the first 25 categories in alphabetic order to make our results more comparable to previous studies \cite{cao2018partial}.

\textbf{VisDA} \cite{visda2017} is another challenging synthetic to real dataset. We use the training set as the synthetic source and the testing set as the real-world target (\textbf{Synthetic$\rightarrow$Real}). Specifically, the training set contains 152K synthetic images generated by rendering 3D models and the testing set contains 72K real images from crops of Youtube Bounding Box dataset \cite{real2017youtube}, both contain 12 categories.  

\subsection{Experimental Setup}
To better study negative transfer effect and evaluate our approach, we need to control the three factors discussed in Section \ref{sec:rethink}, namely \textit{algorithm factor}, \textit{divergence factor} and \textit{target factor}. In our experiments, we adopt the following mechanism to control each of them.

\begin{figure}[ht]
\vskip 0.1in
\begin{center}
  \subfigure[Original]{ 
    \includegraphics[width=.40\columnwidth]{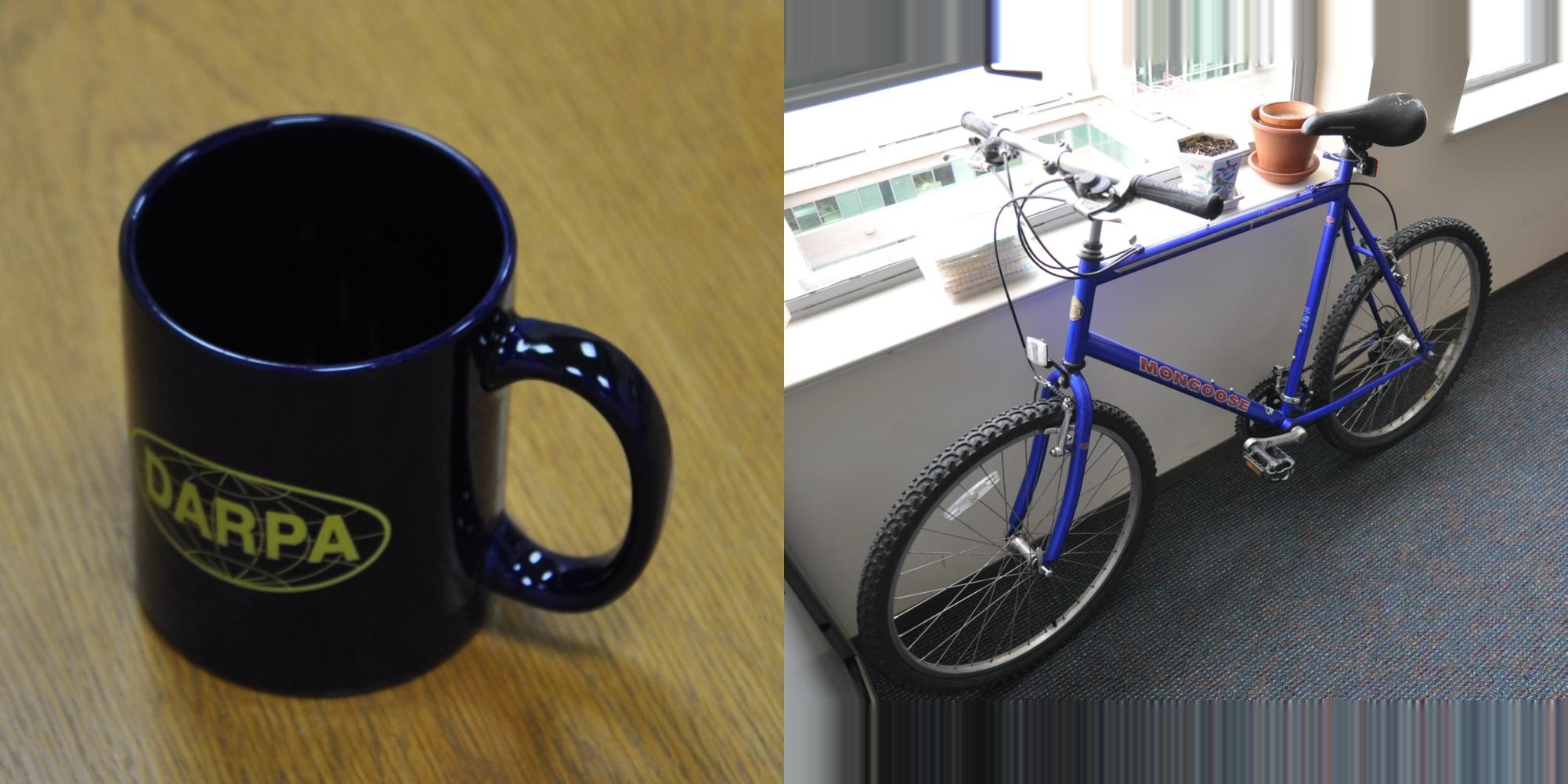} \label{fig:2a} 
  } 
  \subfigure[Perturbed]{% 
    \includegraphics[width=.40\columnwidth]{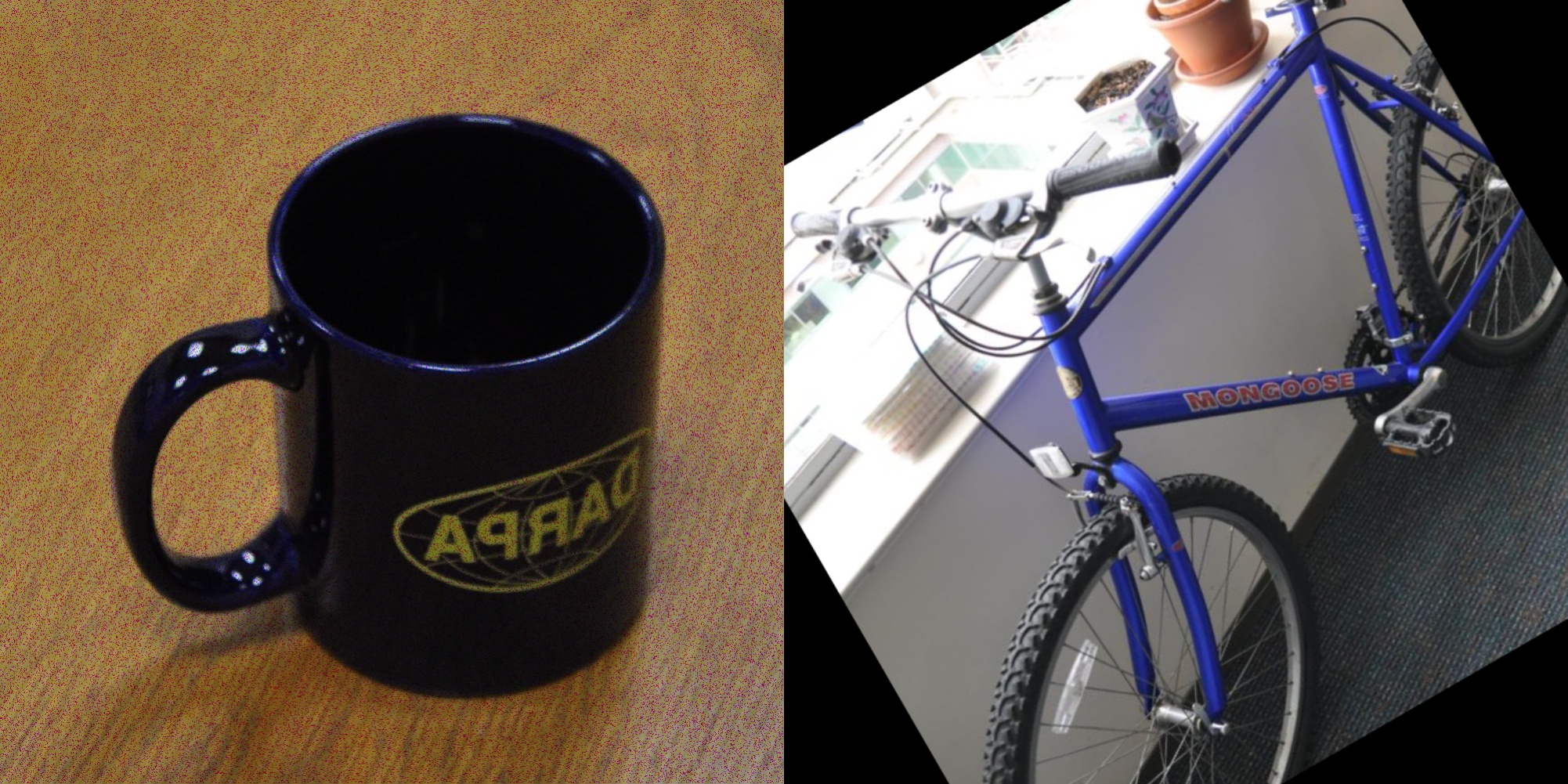} \label{fig:2b} 
  } 
\caption{Example images before \& after perturbation}
\label{fig:2}
\end{center}
\vskip -0.1in
\end{figure}

\textbf{Divergence factor:} Since existing benchmark datasets usually contain domains that are similar to each other, we need to alter their distributions to better observe negative transfer effect. In our experiments, we introduce two perturbation rates $\epsilon_x$ and $\epsilon_y$ to respectively control the marginal divergence and the conditional divergence between two domains. Specifically, for each source domain data we independently draw a Bernoulli variable of probability $\epsilon_x$, and if it returns one, we add a series of random noises to the input image such as random rotation, random salt\&pepper noise, random flipping, etc (examples shown Figure \ref{fig:2}). According to studies in \cite{szegedy2014intriguing,azulay2018deep}, such perturbation is enough to cause misclassification for neural networks and therefore is sufficient for our purpose. In addition, we draw a second independent Bernoulli variable of probability $\epsilon_y$ and assign a randomly picked label if it returns one.

\textbf{Target factor:} Similar to previous works, we use all labeled source data for training. For the target data, we first split 50$\%$ as training set and the rest 50$\%$ for testing. In addition, we use all of target training data as unlabeled target data and use $L_{\%}$ percent of them as labeled target data. A symmetric study of source data can be found in \cite{wang2018towards}.

\begin{table*}[ht]
\caption{\footnotesize Classification accuracy (\%) of DANN and $\mbox{DANN}_{\mbox{gate}}$ on tasks W$\rightarrow$D and A$\rightarrow$D. Perturbation rates are set equal, i.e. $\epsilon=\epsilon_x=\epsilon_y$. $NTG_1$ and $NTG_2$ are negative transfer gaps for DANN and $\mbox{DANN}_{\mbox{gate}}$. $\Delta$ is the performance gain of $\mbox{DANN}_{\mbox{gate}}$ compared to DANN.}
\label{table:single_DANN}
\vskip 0.1in
\begin{center}
\scriptsize
\begin{tabular}{c c c c c c | c c c c c c}
\toprule
\multirow{2}{*}{} & \multicolumn{5}{c}{W$\rightarrow$D} & \multicolumn{5}{c}{A$\rightarrow$D} \\
\cmidrule(lr){2-6} \cmidrule(lr){7-11}
& $\epsilon=$0.0 & $\epsilon=$0.3 & $\epsilon=$0.7 & $\epsilon=$0.9 & Avg & $\epsilon=$0.0 &  $\epsilon=$0.3 & $\epsilon=$0.7 & $\epsilon=$0.9 & Avg &  $L_{\%}$ \\
\bottomrule
DANN & 99.1$\pm$0.8 & 83.2$\pm$1.4 & 47.2$\pm$2.7 & 32.2$\pm$3.5 & 65.4 & 76.2$\pm$1.5 & 40.9$\pm$1.1 & 21.3$\pm$2.7 & 12.9$\pm$3.7 & 37.8 & \multirow{5}{*}{0\%} \\
$\mbox{NTG}_1$ & -96.5 & -80.3 & -44.1 & -28.3 & -62.3 & -73.7 & -37.3 & -17.2 & -9.7 & -34.5 & \\
$\mbox{DANN}_{\mbox{gate}}$ & 98.9$\pm$0.6 & 83.3$\pm$2.1 & 48.4$\pm$2.5 & 32.1$\pm$3.1 & 65.7 & 76.0$\pm$1.2 & 41.0$\pm$1.6 & 21.5$\pm$3.1 & 13.2$\pm$2.4 & 37.9 & \\
$\mbox{NTG}_2$ & -96.3 & -80.4 & -45.3 & -28.2 & -62.6 & -73.5 & -37.4 & -17.4 & -10.0 & -34.6 & \\
$\Delta$ & \color{bured}{$\downarrow$0.2} & \color{dgreen}{$\uparrow$0.1} & \color{dgreen}{$\uparrow$1.2} & \color{bured}{$\downarrow$0.1} & \color{dgreen}{$\uparrow$0.3} & \color{bured}{$\downarrow$0.2} & \color{dgreen}{$\uparrow$0.1} & \color{dgreen}{$\uparrow$0.2} & \color{dgreen}{$\uparrow$0.3} & \color{dgreen}{$\uparrow$0.1} & \\
\Xhline{\arrayrulewidth}
DANN & 99.5$\pm$0.4 & 86.8$\pm$2.8 & 73.1$\pm$3.3 & 48.8$\pm$4.3 & 77.0 & 78.6$\pm$2.7 & 54.8$\pm$3.1 & 49.6$\pm$2.1 & 32.3$\pm$2.6 & 53.8 & \multirow{5}{*}{10\%} \\
$\mbox{NTG}_1$ & -48.7 & -37.8 & -23.6 & 1.6 & -27.1 & -28.4 & -4.4 & 1.2 & 18.4 & -3.3 &\\
$\mbox{DANN}_{\mbox{gate}}$ & 99.2$\pm$0.3 & 85.4$\pm$2.6 & 79.4$\pm$2.9 & 50.4$\pm$3.2 & 78.6 & 85.1$\pm$1.7 & 60.2$\pm$2.1 & 58.3$\pm$2.0 & 49.1$\pm$2.5 & 63.2 & \\
$\mbox{NTG}_2$  & -48.4 & -36.4 & -29.9 & 0.0 & -28.7 & -34.9 & -9.8 & -7.5 & 1.6 & -12.7 &  \\
$\Delta$ & \color{bured}{$\downarrow$0.3} & \color{bured}{$\downarrow$1.4} & \color{dgreen}{$\uparrow$6.3} & \color{dgreen}{$\uparrow$1.6} & \color{dgreen}{$\uparrow$1.6} & \color{dgreen}{$\uparrow$6.5} & \color{dgreen}{$\uparrow$5.4} & \color{dgreen}{$\uparrow$8.7} & \color{dgreen}{$\uparrow$16.8} & \color{dgreen}{$\uparrow$9.4} &  \\
\Xhline{\arrayrulewidth}
DANN & 99.6$\pm$0.2 & 89.7$\pm$1.6 & 78.4$\pm$2.5 & 70.5$\pm$4.3 & 84.6 &  80.2$\pm$2.0 & 73.3$\pm$2.2 & 70.2$\pm$3.3 & 51.3$\pm$4.3 & 68.8 & \multirow{5}{*}{30\%} \\
$\mbox{NTG}_1$ & -18.5 & -10.3 & 1.8 & 8.2 & -4.7 & -1.5 & 6.5 & 8.9 & 28.4 & 10.6 & \\
$\mbox{DANN}_{\mbox{gate}}$ & 100.0$\pm$0.1 & 90.4$\pm$1.8 & 82.0$\pm$1.8 & 79.9$\pm$3.8 & 88.1 & 89.0$\pm$1.5 & 82.6$\pm$1.0 & 81.3$\pm$2.1 & 80.6$\pm$1.8 & 83.4 & \\
$\mbox{NTG}_2$ & -18.9 & -11.0 & -1.8 & -1.2 & -8.2 & -10.3 & -2.8 & -2.2 & -0.9 & -4.1 & \\
$\Delta$ & \color{dgreen}{$\uparrow$0.4} & \color{dgreen}{$\uparrow$0.7} & \color{dgreen}{$\uparrow$3.6} & \color{dgreen}{$\uparrow$9.4} & \color{dgreen}{$\uparrow$2.6} & \color{dgreen}{$\uparrow$8.8} & \color{dgreen}{$\uparrow$9.3} & \color{dgreen}{$\uparrow$11.1} & \color{dgreen}{$\uparrow$29.3} & \color{dgreen}{$\uparrow$14.6} & \\
\Xhline{\arrayrulewidth}
DANN & 100.0$\pm$0.0 & 92.2$\pm$1.7 & 85.8$\pm$2.3 & 78.2$\pm$4.8 & 89.1 & 84.5$\pm$1.9 & 77.6$\pm$3.8 & 70.6$\pm$4.9 & 65.4$\pm$6.3 & 74.5 & \multirow{5}{*}{50\%} \\
$\mbox{NTG}_1$ & -11.7 & -3.2 & 3.8 & 10.4 & -0.2 & 4.6 & 12.1 & 18.8 & 23.2 & 14.7 & \\
$\mbox{DANN}_{\mbox{gate}}$ & 100.0$\pm$0.0 & 93.3$\pm$1.7 & 91.2$\pm$1.5 & 89.5$\pm$3.4 & 92.5 & 93.2$\pm$1.3 & 91.4$\pm$1.2 & 90.2$\pm$2.0 & 89.8$\pm$1.9 & 91.2 & \\
$\mbox{NTG}_2$ & -11.7 & -4.3 & -1.6 & -0.9 & -4.6 & -4.1 & -1.7 & -0.8 & -1.2 & -2.0 & \\
$\Delta$ & $\rightarrow$0.0 & \color{dgreen}{$\uparrow$1.1} & \color{dgreen}{$\uparrow$5.4} & \color{dgreen}{$\uparrow$11.3} & \color{dgreen}{$\uparrow$4.5} & \color{dgreen}{$\uparrow$8.7} & \color{dgreen}{$\uparrow$13.8} & \color{dgreen}{$\uparrow$19.6} & \color{dgreen}{$\uparrow$24.4} & \color{dgreen}{$\uparrow$16.7} &\\
\bottomrule
\end{tabular}
\end{center}
\vskip -0.1in
\end{table*}

\textbf{Algorithm factor:} To provide a more comprehensive study of negative transfer, we evaluate the performance of eight transfer learning methods of five categories: \textbf{TCA} \cite{pan2011domain}, \textbf{KMM} \cite{huang2007correcting}, \textbf{DAN} \cite{long2015learning}, \textbf{DCORAL} \cite{sun2016deep}, \textbf{DANN} a.k.a RevGrad \cite{ganin2015unsupervised}, \textbf{ADDA} \cite{tzeng2015simultaneous}, \textbf{PADA} \cite{cao2018partial}, \textbf{GTA} \cite{sankaranarayanan2017generate}. Specifically, (1) TCA is a conventional method based on MMD-regularized PCA, (2) KMM is a conventional sample reweighting method, (3) DAN and DCORAL are non-adversarial deep methods which use a distribution measurement as an extra loss, (4) DANN, ADDA and PADA use adversarial learning and directly train a discriminator, (5) GTA is a GAN based method that includes a generator to generate actual images in additional to the discriminator. We mainly follow the default settings and training procedures for model selection as explained in their respective papers. However, for fair comparison, we use the same feature extractor and classifier architecture for all deep methods. In particular, we use a modified LeNet as detailed in \cite{sankaranarayanan2017generate} for the Digits dataset. For other datasets, we fine-tune from the ResNet-50 \cite{he2016deep} pretrained on ImageNet with an added 256-dimension bottleneck layer between the \textit{res5c} and \textit{fc} layers. To compare the performance of our proposed approach, we adapt a gated version for each of the six deep methods (e.g \textbf{$\mbox{DANN}_{\mbox{gate}}$} is the gated DANN). Specifically, we extend DANN, ADDA and PADA straightforwardly as described in Section 4.2. For GTA, we extend the discriminator to take in class labels and output domain label predictions as gates. For DAN and DCORAL, we add an extra discriminator network to be used as gates but the general network is not trained adversarially. For hyper-parameters, we set $\lambda=1$ and $\mu$ progressively increased from 0 to 1 in all our experiments. For each transfer task, we compare the average classification accuracy over five random repeats. To test whether negative transfer occurs, we measure the negative transfer gap (NTG) as the gap between the accuracy of target-only baseline and that of the original method. For instance, for DANN, the target-only baseline is $\mbox{DANN}_T$ which treats labeled target data as ``source" data and uses unlabeled data as usual. A positive NTG indicates the occurrence of negative transfer and vice versa.

\begin{table*}[ht]
\caption{\footnotesize Classification accuracy (\%) of state-of-the-art methods on four benchmark datasets with negative transfer gap shown in brackets. Perturbation rates are fixed at $\epsilon_x=\epsilon_y=0.7$. Target labeled ratio is set at $L_{\%}=10\%$ and we further enforce each task to use at most 3 labeled target samples per class.}
\label{table:main}
\vskip 0.1in
\begin{center}
\scriptsize
\begin{tabular}{c c c c c c c c c c c}
\toprule
 & Digits  & \multicolumn{3}{c}{Office-31} & \multicolumn{3}{c}{Office-Home} & VisDA & \\
\cmidrule(lr){2-2} \cmidrule(lr){3-5} \cmidrule(lr){6-8} \cmidrule(lr){9-9} 
Method & SVHN$\rightarrow$MNIST & W$\rightarrow$D & A$\rightarrow$D & D$\rightarrow$A & Ar$\rightarrow$Rw & Cl$\rightarrow$Rw & Pr$\rightarrow$Rw & Synthetic$\rightarrow$Real & Avg \\
\bottomrule
TCA\cite{pan2011domain} & 58.7(18.2) & 54.2(-4.2) & 11.4(20.5) & 13.1(18.4) & - & - & - & - & 34.4(13.2)\\
KMM\cite{huang2007correcting} & 70.9(6.0) & 58.7(-8.5) & 18.5(13.4) & 17.7(13.8) & - & - & - & - & 41.5(6.2)\\
\Xhline{\arrayrulewidth}
DAN\cite{long2015learning} & 78.5(-4.4) & 76.3(-19.5) & 55.0(-1.3) & 39.2(4.9) & 43.2(3.8) & 30.2(5.8) & 47.2(4.0) & 28.4(7.2) & 49.8(0.1)\\
$\mbox{DAN}_{\mbox{gate}}$ & 82.2(-8.1) & 78.7(-21.9) & 60.4(-6.7) & 43.9(0.2) & 46.8(0.2) & 38.0(-2.0) & 50.4(0.8) & 36.2(-0.6) & 54.6(-4.7)\\
$\Delta_{\mbox{DAN}}$ & \color{dgreen}{$\uparrow$3.7} & \color{dgreen}{$\uparrow$2.4} & \color{dgreen}{$\uparrow$5.4} & \color{dgreen}{$\uparrow$4.7} & \color{dgreen}{$\uparrow$3.6} & \color{dgreen}{$\uparrow$7.8} & \color{dgreen}{$\uparrow$3.2} & \color{dgreen}{$\uparrow$7.8} & \color{dgreen}{$\uparrow$4.8}\\
DCORAL\cite{sun2016deep} & 75.2(-1.2) & 75.7(-18.9) & 53.8(-0.4) & 37.4(5.0) & 44.0(3.7) & 32.4(4.1) & 48.0(2.2) & 30.5(5.7) & 49.6(0.0)\\
$\mbox{DCORAL}_{\mbox{gate}}$ & 81.0(-7.0) & 78.2(-21.4) & 59.0(-5.6) & 43.2(-0.8) & 48.5(-0.8) & 40.0(-3.5) & 51.6(-1.4) & 35.8(0.4) & 54.7(-5.1) \\
$\Delta_{\mbox{DCORAL}}$ & \color{dgreen}{$\uparrow$5.8} & \color{dgreen}{$\uparrow$2.5} & \color{dgreen}{$\uparrow$5.2} & \color{dgreen}{$\uparrow$5.8} & \color{dgreen}{$\uparrow$4.5} & \color{dgreen}{$\uparrow$7.6} & \color{dgreen}{$\uparrow$3.6} & \color{dgreen}{$\uparrow$5.3} & \color{dgreen}{$\uparrow$5.1}\\
DANN\cite{ganin2015unsupervised} & 68.3(7.7) & 75.0(-19.2) & 51.0(2.3) & 38.2(5.6) & 42.8(4.2) & 28.5(7.7) & 42.0(10.0) & 29.9(6.0) & 47.0(3.0)\\
$\mbox{DANN}_{\mbox{gate}}$ & 78.1(-2.1) & 80.2(-24.4) & 61.8(-8.5) & 48.3(-4.5) & 51.2(-4.2) & 43.8(-7.6) & 55.2(-3.2) & 40.5(-4.6) & 57.4(-7.4)\\
$\Delta_{\mbox{DANN}}$ & \color{dgreen}{$\uparrow$9.8} & \color{dgreen}{$\uparrow$5.2} & \color{dgreen}{$\uparrow$10.8} & \color{dgreen}{$\uparrow$10.1} & \color{dgreen}{$\uparrow$9.4} & \color{dgreen}{$\uparrow$14.7} & \color{dgreen}{$\uparrow$13.2} & \color{dgreen}{$\uparrow$10.6} & \color{dgreen}{$\uparrow$10.4}\\
ADDA\cite{tzeng2017adversarial} & 63.2(12.2) & 74.5(-18.1) & 49.9(2.2) & 38.3(5.1) & 41.4(6.0) & 25.2(13.5) & 43.2(7.2) & 28.0(7.3) & 45.5(4.4)\\
$\mbox{ADDA}_{\mbox{gate}}$ & 79.4(-4.0) & 82.9(-26.5) & 64.2(-12.1) & 47.7(-4.3) & 52.2(-4.8) & 48.0(-9.3) & 58.2(-7.8) & 43.0(-7.7) & 59.5(-9.6)\\
$\Delta_{\mbox{ADDA}}$ & \color{dgreen}{$\uparrow$16.2} & \color{dgreen}{$\uparrow$8.4} & \color{dgreen}{$\uparrow$14.3} & \color{dgreen}{$\uparrow$9.4} & \color{dgreen}{$\uparrow$10.8} & \color{dgreen}{$\uparrow$22.8} & \color{dgreen}{$\uparrow$15.0} & \color{dgreen}{$\uparrow$15.0} & \color{dgreen}{$\uparrow$14.0}\\
PADA\cite{cao2018partial} & 69.7(6.5) & 75.5(-19.0) & 50.2(1.9) & 38.7(5.1) & 43.2(3.8) & 30.1(5.5) & 43.4(6.6) & 32.2(5.5) & 47.9(2.0)\\
$\mbox{PADA}_{\mbox{gate}}$ & 81.8(-5.6) & 81.6(-25.1) & 62.1(-10.0) & 44.8(-1.0) & 52.8(-5.8) & 45.2(-9.6) & 54.5(-4.5) & 41.4(-5.7) & 58.0(-8.1) \\
$\Delta_{\mbox{PADA}}$ & \color{dgreen}{$\uparrow$12.1} & \color{dgreen}{$\uparrow$5.9} & \color{dgreen}{$\uparrow$11.9} & \color{dgreen}{$\uparrow$6.1} & \color{dgreen}{$\uparrow$9.6} & \color{dgreen}{$\uparrow$15.1} & \color{dgreen}{$\uparrow$11.1} & \color{dgreen}{$\uparrow$11.2} & \color{dgreen}{$\uparrow$10.1}\\
GTA\cite{sankaranarayanan2017generate} & 81.2(-6.8) & 78.9(-20.5) & 58.4(-7.2) & 42.2(2.8) & 48.2(1.0) & 33.1(5.1) & 50.2(-0.1) & 31.2(4.2) & 52.9(-2.7)\\
$\mbox{GTA}_{\mbox{gate}}$ & 83.3(-8.9) & 85.8(-27.4) & 66.7(-15.5) & 48.5(-3.5) & 55.0(-5.8) & 44.9(-6.7) & 58.0(-7.7) & 43.8(-8.4) & 60.8(-10.6)\\
$\Delta_{\mbox{GTA}}$ & \color{dgreen}{$\uparrow$2.1} & \color{dgreen}{$\uparrow$6.9} & \color{dgreen}{$\uparrow$8.3} & \color{dgreen}{$\uparrow$6.3} & \color{dgreen}{$\uparrow$6.8} & \color{dgreen}{$\uparrow$11.8} & \color{dgreen}{$\uparrow$7.8} & \color{dgreen}{$\uparrow$12.6} & \color{dgreen}{$\uparrow$7.9}\\
\Xhline{\arrayrulewidth}
$\Delta$Avg & \color{dgreen}{$\uparrow$8.3} & \color{dgreen}{$\uparrow$5.2} & \color{dgreen}{$\uparrow$8.1} & \color{dgreen}{$\uparrow$7.1} & \color{dgreen}{$\uparrow$7.5} & \color{dgreen}{$\uparrow$13.3} & \color{dgreen}{$\uparrow$8.9} & \color{dgreen}{$\uparrow$10.4} & \\
\bottomrule
\end{tabular}
\end{center}
\vskip -0.1in
\end{table*}

\begin{figure}[t!]
\vskip 0.1in
\begin{center}
  \subfigure[$L_{\%}$ fixed at 20\%]{ 
    \includegraphics[width=.44\columnwidth]{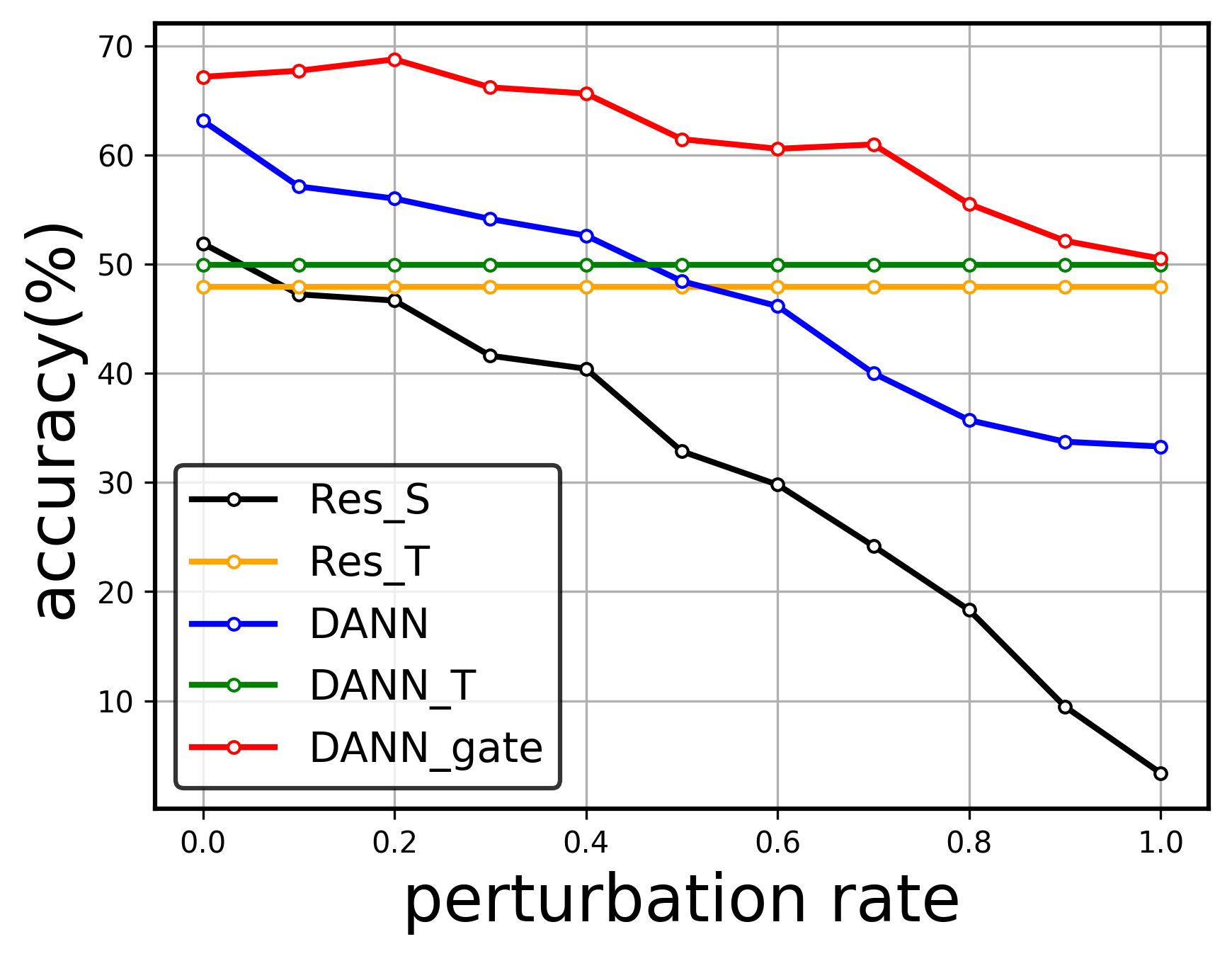} \label{fig:3a} 
  } 
  \hfill
  \subfigure[$\epsilon$ fixed at 0.2]{% 
    \includegraphics[width=.44\columnwidth]{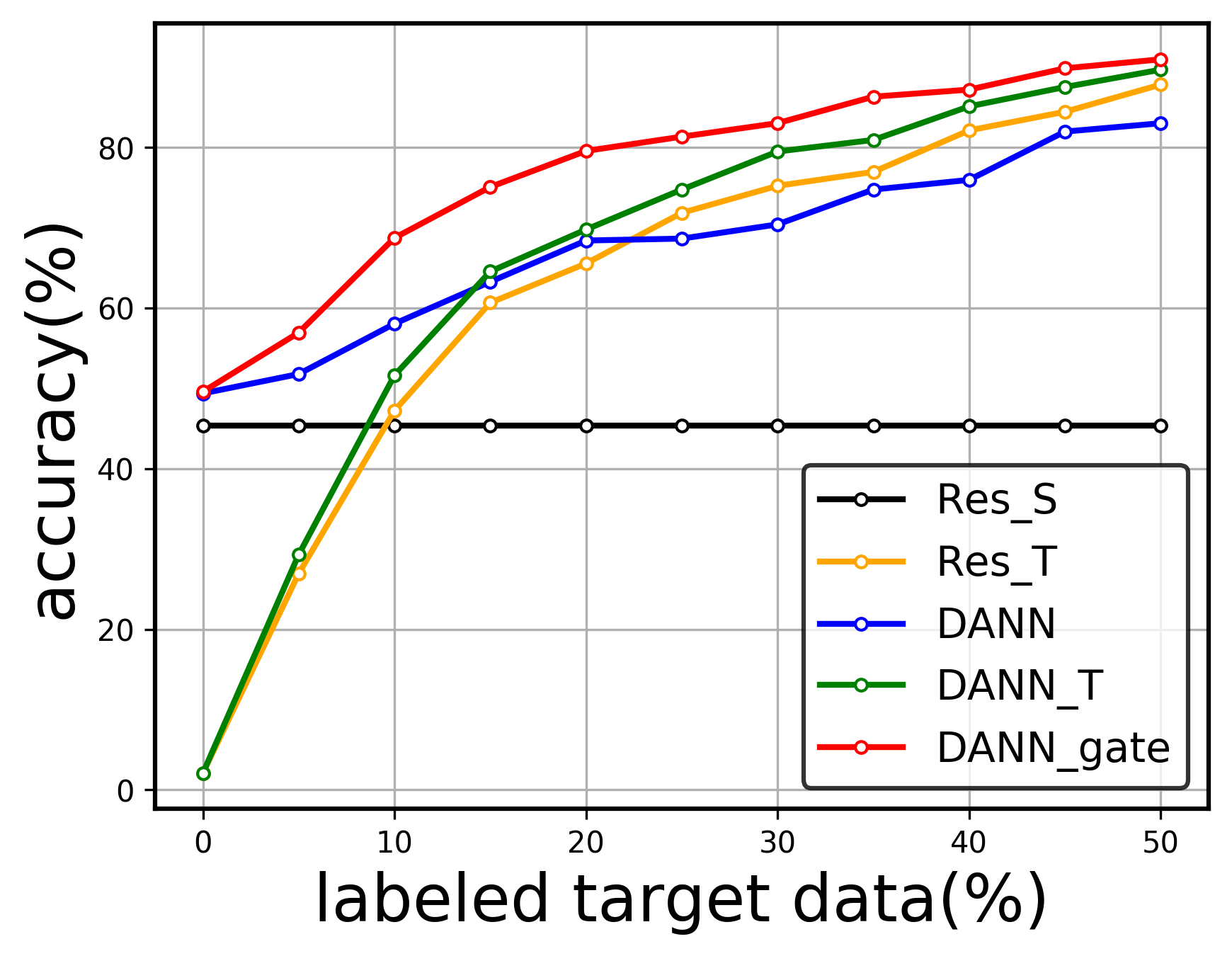} \label{fig:3b} 
  } 
\caption{\footnotesize Incremental performance on task Pr$\rightarrow$Rw. $\mbox{Res}_S$ and $\mbox{Res}_T$ are ResNet-50 baselines trained using only source data and only target data. Perturbation rates are set equal, i.e. $\epsilon=\epsilon_x=\epsilon_y$.}
\label{fig:increment}
\end{center}
\vskip -0.4in
\end{figure}

\subsection{Results and Analysis}
\subsubsection{Study of Negative Transfer}
To reveal the three dependent factors, we study the effect of negative transfer under different methods with varying perturbation rates ($\epsilon_x, \epsilon_y$) and target labeled data ($L_{\%}$).

\textbf{Divergence factor.} The performance of DANN under different settings of $\epsilon$ and $L_{\%}$ on two tasks of Office-31 are shown in Table \ref{table:single_DANN}. We observe an increasing negative transfer gap as we increase the perturbation rate in all cases. In some cases such as $L_{\%} = 10\%$, we can even observe a change in the sign of NTG. For a more fine-grained study, we investigate a wider spectrum of distribution divergence by gradually increasing $\epsilon$ from 0.0 to 1.0 in Figure \ref{fig:3a}. Although DANN is better than $\mbox{DANN}_T$ when $\epsilon$ is small, its performance degrades quickly as $\epsilon$ increases and drops below $\mbox{DANN}_T$, indicating the occurrence of negative transfer. On the other hand, by fixing $\epsilon_y=0$ and using two domains \textbf{W} and \textbf{D} that are known to be particularly similar, we study negative transfer under the assumption of covariate shift in Table \ref{table：covariate shift}, and observe that negative transfer does \textit{not} occur even with high $\epsilon_x$ and descent $L_{\%}$. These experimental results confirms that the distribution divergence is an important factor of negative transfer.

\begin{table}[H]
\caption{\footnotesize Classification accuracy (\%) under the Covariate Shift assumption on task W$\rightarrow$D. $\epsilon_y$ is fixed at 0. Negative transfer gap is shown in brackets.}
\label{table：covariate shift}
\begin{center}
\footnotesize
\begin{tabular}{c | c c}
\toprule
%  & \multicolumn{2}{c}{Setting} \\
% \cmidrule(lr){2-3} 
Method & $\epsilon_x$=0.7 $L_{\%}$=10\% & $\epsilon_x$=1.0 $L_{\%}$=30\%\\
\bottomrule
DAN & 81.2(-29.3) & 85.8(-6.2)\\
DANN & 83.0(-30.8) & 86.1(-6.5)\\
GTA & 85.5(-33.5) & 88.1(-8.0)\\
\bottomrule
\end{tabular}
\end{center}
\vspace{-2em}
\end{table}
\textbf{Target factor.} Fixing a specific $\epsilon$, we observe that the negative transfer gap increases as $L_{\%}$ increases in Table \ref{table:single_DANN}. In the extreme case of unsupervised adaptation ($L_{\%}=0\%$), NTG stays negative even if two domains are far apart ($\epsilon = 0.9$). In Figure \ref{fig:3b}, we fix $\epsilon=0.2$ and plot the performance curve as $L_{\%}$ increases. We can see that while both DANN and $\mbox{DANN}_T$ perform better with more labeled target data, DANN is affected by the divergence factor and outperformed by $\mbox{DANN}_T$ when $L_{\%}$ becomes larger. This observation shows that negative transfer is relative and it depends on target labeled data.

\textbf{Algorithm factor.} In Table \ref{table:main}, we compare the results of all methods under a more practically interesting scenario of moderately different distributions and limited amount of labeled target data. We observe that some methods are more vulnerable to negative transfer then the other even using the same training data. For conventional methods, instance-reweighting method KMM achieves smaller NTG compared to feature selection method TCA, possibly because KMM can assign small weights to source instances with dissimilar input features. For deep methods, we find GTA to be the most robust method against negative transfer since it takes both label information and random noises as inputs to the generator network. More interestingly, we observe that methods based on distribution measurement such as MMD (e.g. DAN) achieve smaller NTG than methods based on adversarial networks (e.g. DANN), even though the later tends to perform better when distributions are similar. This is consistent with findings in previous works \cite{cao2017partial} and one possible explanation is that adversarial network's better capability of matching source and target domains leads to more severe negative transfer. Similarly, ADDA has better matching power by using two separate feature extractors, but it results in larger NTG compared to DANN.
\begin{figure*}[ht]
\vskip -0.1in
\begin{center}
  \subfigure[DANN]{ 
    \includegraphics[width=.48\columnwidth]{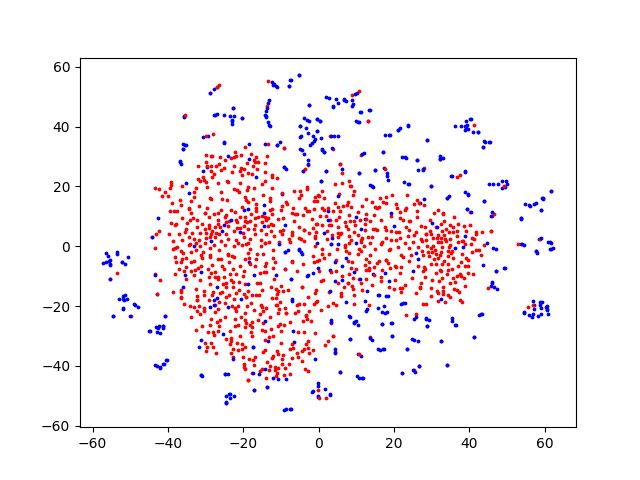} \label{fig:4a} 
  } 
  \hfill
  \subfigure[$\mbox{DANN}_{\mbox{gate}}$]{% 
    \includegraphics[width=.48\columnwidth]{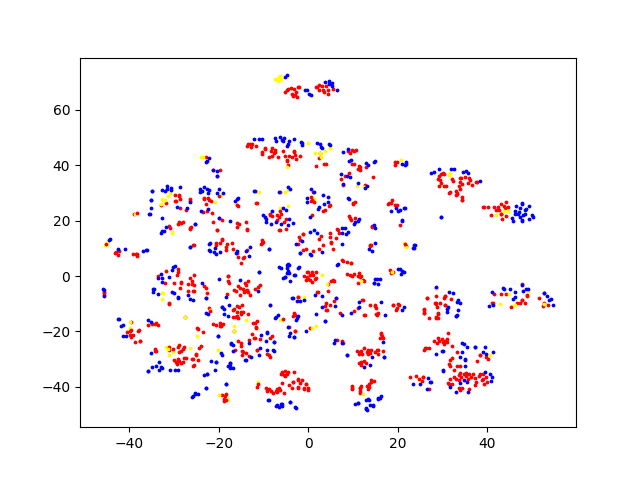} \label{fig:4b} 
  }
  \hfill 
  \subfigure[$\mbox{DANN}_{\mbox{gate}}$ (source data with large weights)]{% 
    \includegraphics[width=.48\columnwidth]{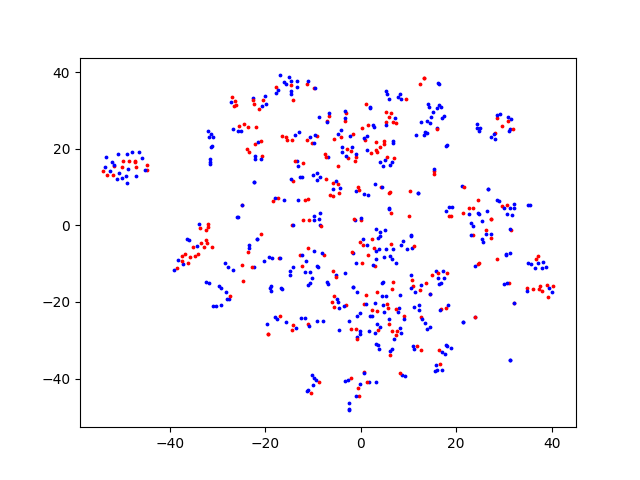} \label{fig:4c} 
  } 
  \hfill
  \rulesep 
  \hfill
  \subfigure[Source Sample Weights]{% 
    \includegraphics[width=.48\columnwidth]{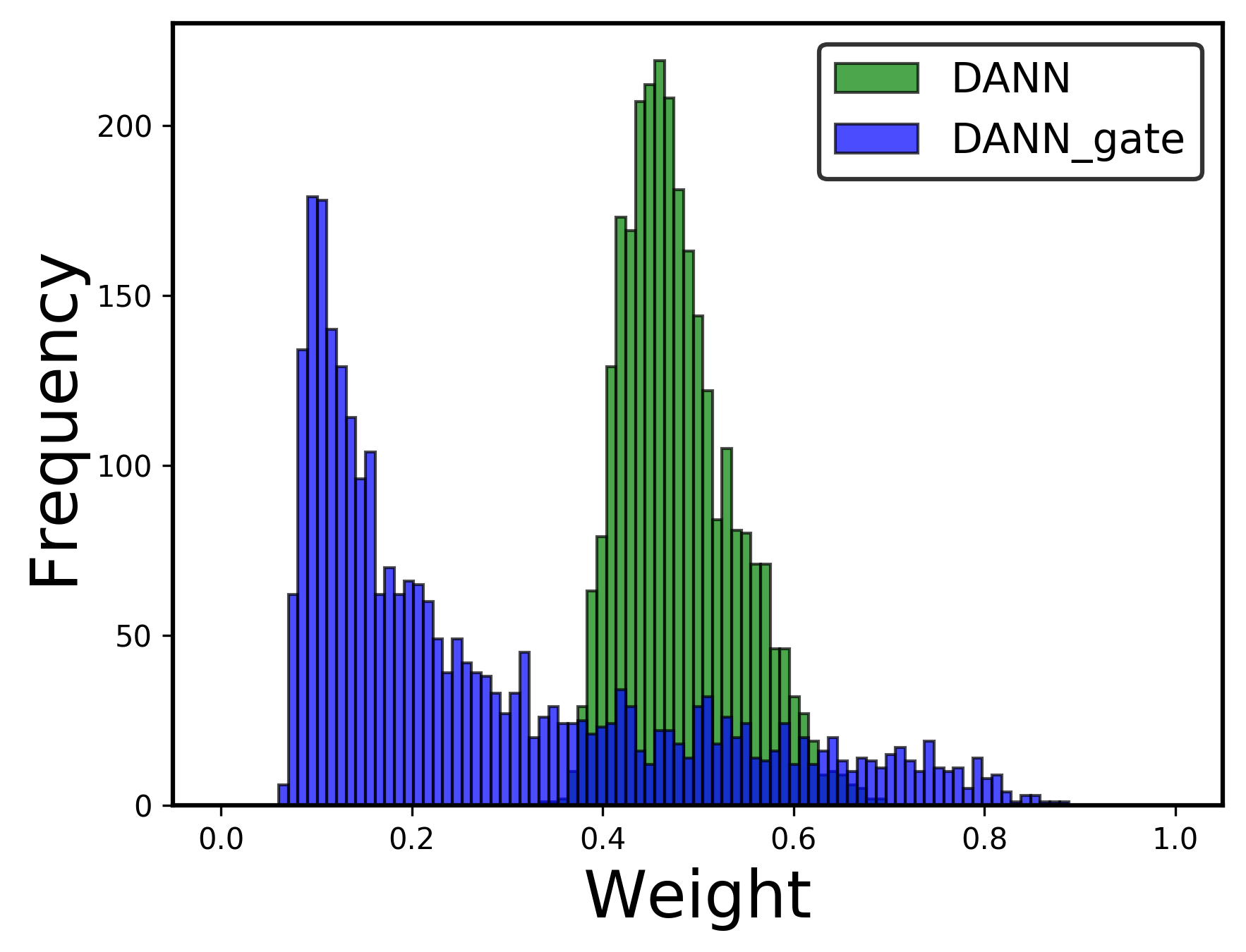} \label{fig:4d} 
  }
  \subfigure[DANN]{ 
    \includegraphics[width=.48\columnwidth]{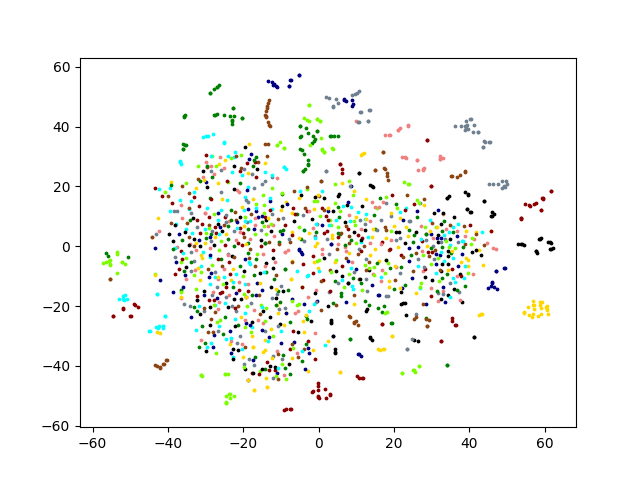} \label{fig:4e} 
  } 
  \hfill
  \subfigure[$\mbox{DANN}_{\mbox{gate}}$]{% 
    \includegraphics[width=.48\columnwidth]{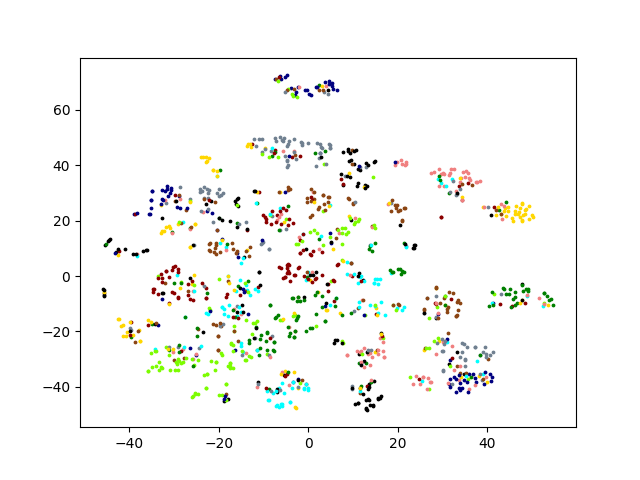} \label{fig:4f} 
  }
  \hfill 
  \subfigure[$\mbox{DANN}_{\mbox{gate}}$ (source data with large weights)]{% 
    \includegraphics[width=.48\columnwidth]{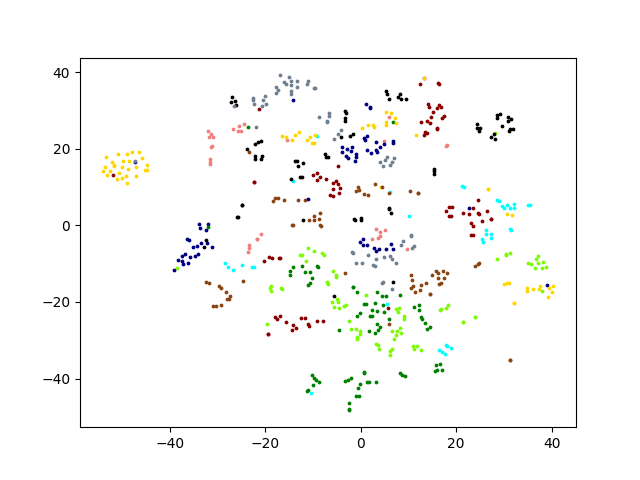} \label{fig:4g} 
  }
  \hfill
  \rulesep
  \hfill
  \subfigure[Left: DANN Right:$\mbox{DANN}_{\mbox{gate}}$]{% 
    \includegraphics[width=.48\columnwidth]{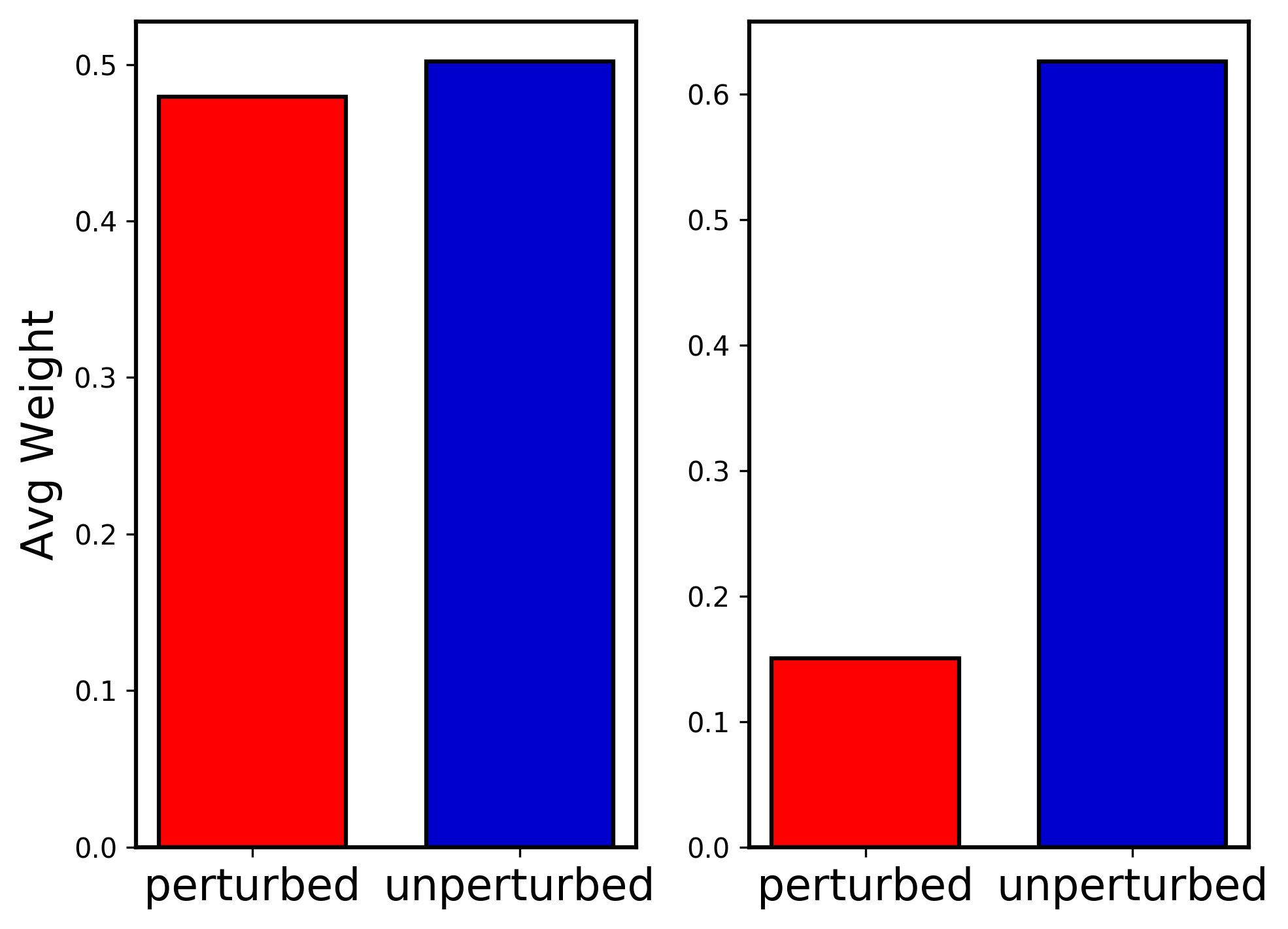} \label{fig:4h} 
  } 
\caption{Visualization on A$\rightarrow$W, with $\epsilon=0.7$, $L_{\%}=30\%$. \textbf{Left:} The t-SNE visualization. First row shows domain info with red for source samples (yellow for weights $>$ 0.4) and blue for target samples. Second tow shows corresponding class info. \textbf{Right:} Top shows the histogram of discriminator weights for source samples. Bottom shows average weights for perturbed and unperturbed samples.}
\label{fig:tsne}
\end{center}
\vspace{-2em}
\end{figure*}

\begin{table}[ht]
\vspace{-0.1in}
\caption{Ablation Study on task A$\rightarrow$D. $\mbox{DANN}_{\mbox{gate-only}}$ applies only the discriminator gate while $\mbox{DANN}_{\mbox{label-only}}$ only uses label information without the gate. $\mbox{DANN}_{\mbox{joint}}$ is a variant of $\mbox{DANN}_{\mbox{gate}}$ where the feature network only matches the joint distribution (last two lines of Eq.\ref{eq:aug_adv_obj}), $\mbox{DANN}_{\mbox{marginal}}$ only matches the marginal distribution, %(second line of Eq.\ref{eq:LD}), 
and $\mbox{DANN}_{\mbox{none}}$ matches none of them. $\mbox{DANN}_{\mbox{oracle}}$ excludes perturbed source data via human oracle.}
\label{table:ablation}
\begin{center}
\begin{scriptsize}
\begin{tabular}{l c c c c c}
\toprule
\multirow{2}{*}{} & \multicolumn{4}{c}{Setting ($\epsilon$,$L_{\%}$)} \\
\cmidrule(lr){2-5} 
Method & 0.7, 30\% & 0.7, 10\% & 0.3, 30\% & 0.3, 10\% & Avg \\
\bottomrule
DANN & 70.4 & 49.4 & 72.5 & 54.3 & 61.7 \\
$\mbox{DANN}_T$  & 79.5 & 50.7 & 80.3 & 50.1 & 65.2  \\
$\mbox{DANN}_{\mbox{oracle}}$ & 81.6 & 58.5 & \bf 89.1 & \bf 
85.4 & \bf 78.7\\
\Xhline{\arrayrulewidth}
$\mbox{DANN}_{\mbox{gate-only}}$ & 76.3 & 53.8 & 78.0 & 55.7 & 66.0  \\
$\mbox{DANN}_{\mbox{label-only}}$ & 74.4 & 52.5 & 77.5 & 55.0 & 64.9  \\
$\mbox{DANN}_{\mbox{joint}}$ & 82.3 & 57.6 & 83.1 & 59.4 & 70.6  \\
$\mbox{DANN}_{\mbox{marginal}}$ & 80.6 & 56.5 & 81.5 & 58.6 & 69.3 \\
$\mbox{DANN}_{\mbox{none}}$ & 79.6 & 52.4 & 79.7 & 57.5 & 67.3 \\
$\mbox{DANN}_{\mbox{gate}}$ & \bf 82.5 & \bf 58.7 & 82.7 & 60.7 & 71.2  \\
\bottomrule
\end{tabular}
\end{scriptsize}
\end{center}
\vspace{-2em}
\end{table}

\subsubsection{Evaluation of Discriminator Gate}
We compare our gated models with their respective state-of-the-art methods on the benchmarks in Table \ref{table:main}. Even using limited amount of labeled target data, our proposed method consistently improves the performance for all deep methods on all tasks. More importantly, our method can largely eliminate the negative impact of less related source data and avoid negative transfer (e.g. $\mbox{DANN}_{\mbox{gate}}$ achieves negative average NTG while DANN gets positive NTG). Specifically, our method achieves larger accuracy gains on harder tasks such as synthetic to real-world tasks in Office-Home and VisDA. This is mainly because source domains in these tasks tend to contain more unrelated samples. This finding is also consistent with results in Table \ref{table:single_DANN} and Figure \ref{fig:3a} where we can observe larger performance gains as perturbation rates increase. In the extreme case where the source domain is degenerate ($\epsilon=1.0$ in Figure \ref{fig:3a}), the gated model achieves comparable results to those of $\mbox{DANN}_T$. On the other hand, the results of DANN and $\mbox{DANN}_{\mbox{gate}}$ are similar when source domain is closely related to the target ($\epsilon=0.0$ on task W$\rightarrow$D in Table \ref{table:single_DANN}). This indicates that the discriminator gate can control the trade-off between maximal transfer and alleviating negative impact.\\
\textbf{Ablation Study.} We report the results of ablation study in Table \ref{table:ablation} and analyze the effects of several components in our method subject to different settings of transfer tasks. First, both $\mbox{DANN}_{\mbox{gate-only}}$ and $\mbox{DANN}_{\mbox{label-only}}$ perform better than DANN but worse than $\mbox{DANN}_{\mbox{gate}}$, showing that the discriminator gate and estimating joint distributions can both improve performance but their combination yields full performance benefit. Second, $\mbox{DANN}_{\mbox{joint}}$ obtains higher accuracy results than $\mbox{DANN}_{\mbox{marginal}}$ and $\mbox{DANN}_{\mbox{none}}$ since matching joint distributions is the key to avoid negative transfer when both marginal and conditional distributions shift. However, while $\mbox{DANN}_{\mbox{joint}}$ achieves comparable results as $\mbox{DANN}_{\mbox{gate}}$ when $L_{\%}=30\%$, it performs worse than $\mbox{DANN}_{\mbox{gate}}$ when $L_{\%}=10\%$. This shows that utilizing unlabeled target data to match marginal distributions can be beneficial when labeled target data is scarce. Lastly, it is inspiring to see $\mbox{DANN}_{\mbox{gate}}$ outperforms $\mbox{DANN}_{\mbox{oracle}}$ when perturbation rate is high. This is because less unperturbed source data are used for $\mbox{DANN}_{\mbox{oracle}}$ but $\mbox{DANN}_{\mbox{gate}}$ can utilize perturbed source data that contain related information. This further shows the effectiveness of our approach.\\
\textbf{Feature Visualization.} We visualize the t-SNE embeddings \cite{donahue2014decaf} of the bottleneck representations in Figure \ref{fig:tsne}. The first column shows that, when perturbation rate is high, DANN cannot align the two domains well and it fails to discriminate both source and target classes as different classes are mixed together. The second column illustrates the discriminator gate can improve the alignment by assigning less weights to unrelated source data. For instance, we can see some source data from different classes mixed in the yellow cluster at the center right but they get assigned smaller weights. The third column shows the embeddings after we remove source data with small discriminator weights ($<$ 0.4). We can observe that target data are much better clustered compared to that of DANN. These in-depth results demonstrate the efficacy of discriminator gate method.\\
\textbf{Statistics of Instance Weights.} We illustrate the discriminator output ($\frac{P_T(x^i_s, y^i_s)}{P_T(x^i_s, y^i_s) + P_S(x^i_s, y^i_s)}$) for each source data in Figure \ref{fig:4d}. We can observe that DANN fails to discriminate unrelated source data as all weights concentrate around 0.5 in the middle. On the other hand, $\mbox{DANN}_{\mbox{gate}}$ assigns smaller weights to a large portion of source data (since perturbation rate is high) and thus filters out unrelated information. Figure \ref{fig:4h} further shows that DANN assign similar average weights for perturbed and unperturbed source data while $\mbox{DANN}_{\mbox{gate}}$ outputs much smaller values for perturbed data but higher ones for unperturbed data.

\section{Conclusion}
In this work, we analyze the problem of negative transfer and propose a novel discriminator gate technique to avoid it. We show that negative transfer directly relates to specific algorithms, domain divergence and target data. Experiments demonstrate these factors and the efficacy of our method. Our method consistently improves the performance of base methods and largely avoids negative transfer. Understanding negative transfer in more complex transfer tasks and settings should be addressed in a future research.

\appendix
\section{Negative Transfer Definition}
In this section we show how the negative transfer condition (NTC) in Eq.(3) is derived.

The intuitive definition given earlier in Section 3 does not leads to a rigorous definition. There are two key questions that are not clear: (1) Should negative transfer be defined to be algorithm-specific? (2) What is the negative impact being compared with?

First, if negative transfer is completely algorithm-agnostic, then its definition would be independent to which transfer learning algorithm is being used. Mathematically, this may yield the following:
\begin{equation}
    \min_{A} R_{P_T}(A(\mathcal{S},\mathcal{T})) > \min_{A'} R_{P_T}(A'(\emptyset,\mathcal{T})).
\end{equation}
However, it is easy to see that this condition is never satisfied. 
To show this, given source data $\mathcal{S}$ and target data $\mathcal{T}$, consider an algorithm $A_1$ that minimizes the expected risk on the RHS:
\[ A_1 \in \argmin_{A'} R_{P_T}(A'(\emptyset,\mathcal{T})). \]
Then we can always construct a new algorithm $A_1'$ such that $A_1'(\mathcal{S},\mathcal{T})) = A'(\emptyset,\mathcal{T})$, i.e. $A_1'$ always ignores the source data. As a result, we must have:
\begin{align}
\begin{split}
\min_{A} R_{P_T}(A(\mathcal{S},\mathcal{T})) & \leq R_{P_T}(A_1'(\mathcal{S},\mathcal{T}))\\ 
& = R_{P_T}(A_1(\emptyset,\mathcal{T}))\\ 
& = \min_{A'} R_{P_T}(A'(\emptyset,\mathcal{T}))
\end{split}
\end{align}
Therefore, the condition defined in Eq.(12) is never true and we conclude that negative transfer must be algorithm-specific. This answers the first question.

Given the answer, the condition in Eq.(12) could be modified to consider only a specific transfer algorithm $A$, i.e.,
\begin{equation}
\label{eqn:compared_to_min}
R_{P_T}(A(\mathcal{S},\mathcal{T})) > \min_{A'} R_{P_T}(A'(\emptyset,\mathcal{T})).
\end{equation}
However, there are still two problems with this definitions: 
\begin{enumerate}[leftmargin=*,label=(\alph*)]
\item This condition cannot be measured in practice since we cannot evaluate the RHS even at test time;
\item An algorithm that does not utilize any source at all still satisfies the condition, which is counterintuitive. 
For instance, consider a degenerated algorithm $A_2$ such that $A_2(\mathcal{S},\mathcal{T}) = A_2(\emptyset,\mathcal{T})$ and $R_{P_T}(A_2(\emptyset,\mathcal{T})) > \min_{A'} R_{P_T}(A'(\mathcal{S},\mathcal{T}))$. This algorithm does not perform any meaningful transfer from the source, but negative transfer occurs in this case according to Eq. \eqref{eqn:compared_to_min} since:
$$ R_{P_T}(A_2(\mathcal{S},\mathcal{T})) = R_{P_T}(A_2(\emptyset,\mathcal{T})) > \min_{A'} R_{P_T}(A'(\emptyset,\mathcal{T})). $$
\end{enumerate}
Therefore, it is misleading to only compare with the best possible algorithm and we propose the following definition:\vspace{1em}\\
\textbf{Definition 1 (Negative Transfer).} {\it Given a source dataset $\mathcal{S}$, a target dataset $\mathcal{T}$ and a transfer learning algorithm $A$, the negative transfer condition (NTC) is defined as:}
\begin{equation}
R_{P_T}(A(\mathcal{S},\mathcal{T})) > R_{P_T}(A(\emptyset,\mathcal{T})) \geq \min_{A'} R_{P_T}(A'(\emptyset,\mathcal{T})),
\end{equation}
which is exactly Eq.(3) since the ``$\geq$" constraint on the right side is true for any $A$. This definition of NTC resolves the two questions mentioned above. Furthermore, it is consistent with the intuitive definition and is also tractable at test time.

{\small
\bibliographystyle{ieee}
\bibliography{citation}
}

\clearpage

\end{document}